%% file: arxiv.tex
\documentclass[lettersize,journal]{IEEEtran}

\usepackage{cite}
\usepackage{graphicx}
\usepackage{booktabs}
\usepackage{array}
\usepackage{tabularx}
\usepackage{amsmath}
\usepackage{amssymb}
\usepackage{xcolor}
\usepackage{url}
\usepackage{float}
\usepackage[hidelinks]{hyperref}
\usepackage{stfloats}
\usepackage{flushend}


\graphicspath{{fig/}}

\newcolumntype{L}[1]{>{\raggedright\arraybackslash}p{#1}}
\newcolumntype{Y}{>{\centering\arraybackslash}X}
\title{The Verification Gap in Networked Physical AI: \\
A Post-Semantic Communication Framework}

\author{Shunsuke Saruwatari \\
The University of Osaka, Japan}

\begin{document}

\input{main/arxiv26}

\end{document}

%% file: main/arxiv26.tex
\hypersetup{
  pdftitle={The Verification Gap in Networked Physical AI: A Post-Semantic Communication Framework},
  pdfauthor={Shunsuke Saruwatari},
  pdfsubject={Post-Semantic Communication for Networked Physical AI},
  pdfkeywords={post-semantic communication, verification gap, Networked Physical AI, evidence contract, evidence record, authorized finalization, receiver feedback}
}

\maketitle

\begin{abstract}
A task-effective proposal is not yet a justified physical action. In networked
Physical AI, a proposal may be understood while valid, timely,
proposal-bound evidence or the authority required to finalize an action remains
unavailable. We call this mismatch the \emph{verification gap} and introduce a
\emph{Post-Semantic Communication Framework} for the systems interface between
proposal formation and physical execution. The framework begins with
application-declared evidence requirements, represents qualifying observations
as evidence records, validates supporting and conflicting records through one
path, and separates evidence sufficiency from authorized finalization and a
downstream runtime gate. It further distinguishes \emph{evidence transfer},
which can enlarge the record set reachable by a finalizer, from \emph{evidence
coordination}, which can suppress transmission around records already held at
the finalization endpoint. Finite-state framework checks verify that the
evaluator implements the declared distinctions consistently. Under the declared
model, the controlled communication study exposes a finalizer-dependent
asymmetry: sender-finalized Feedback uses evidence transfer to expand evidence
reachability throughout the feasible plotted region, whereas receiver-finalized
Feedback uses coordination to suppress redundant payload until loss, latency,
freshness, and deadline costs shift selection to One-way. Finally, an
episode-level reporting schema defines common denominators for future measured
Physical-AI studies.
\end{abstract}


\begin{figure*}[!b]
\centering
\includegraphics[width=0.98\textwidth]{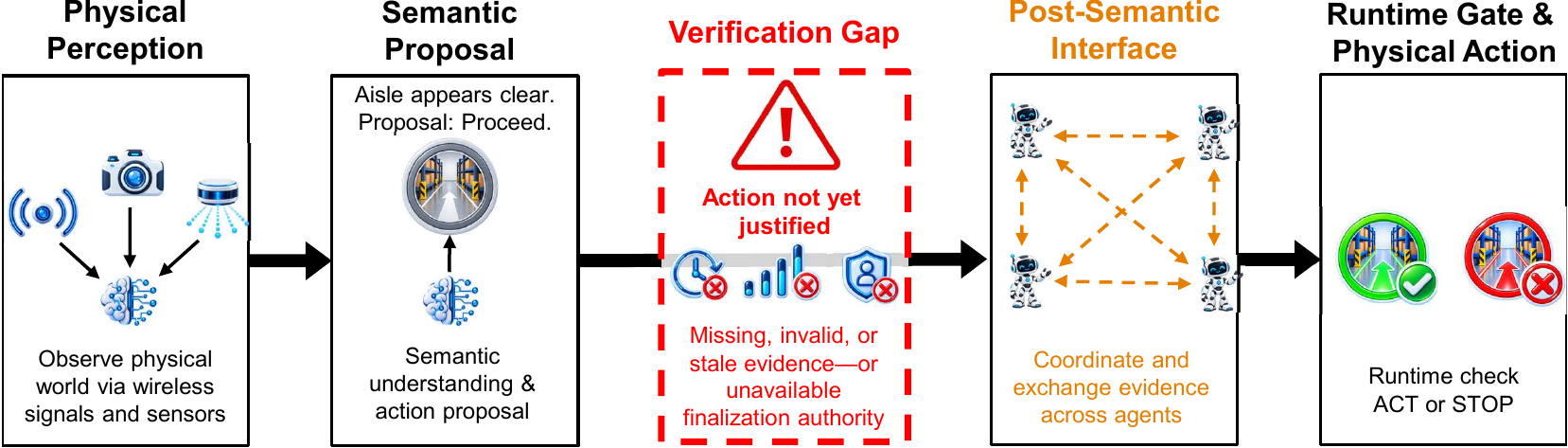}
\caption{The verification gap and post-semantic interface within a networked
Physical-AI loop. Figure~1 locates the interface between semantic proposal
formation and runtime action; Figure~2 defines its internal framework.}
\label{fig:post-semantic-loop}
\end{figure*}

\section{Introduction}

\emph{``The aisle appears clear. Proposal: Proceed.''} A camera-equipped
warehouse robot sends this proposal to another robot approaching an occluded
aisle. The bits arrive, the meaning is decoded, and the proposal may be useful.
The receiving system nevertheless cannot justify motion if the application
requires a fresh Visual-clearance record and a Radio-no-motion record bound to
this proposal, but only one is available. A nearby agent may hold the missing
record; it may instead hold an admissible conflict that should stop the action.
A task-effective proposal is not yet a justified physical action.

The same distinction arises when a cooperative vehicle receives a trajectory,
a manipulator receives a grasp proposal, or an inspection drone receives a
route. Networked Physical AI inherits long-standing cyber--physical and cloud-
robotics concerns: sensing, computation, communication, and actuation are
distributed across components and time \cite{rajkumarcps,kehoecloud}. This
distribution can also separate evidence location from finalization authority.
We define the \emph{verification gap} as the mismatch between a task-effective
proposal and the valid, timely, proposal-bound evidence and authorization
required at the permitted finalizer before action. Figure~\ref{fig:post-semantic-loop}
places this gap inside a Networked Physical-AI loop
\cite{physicalaisurvey,wmcdt}.

\emph{Post-semantic communication} names the systems interface after semantic
or task-level proposal formation and before physical execution. At this
interface, agents expose evidence holdings, transfer validated evidence
records, re-evaluate application-declared requirements, and make a sufficient
record set available to the authorized finalizer. The interface does not
prescribe a transport, topology, learned representation, or number of agents.
It identifies a responsibility boundary that is easily hidden inside a generic
``receiver decision.'' The downstream runtime gate remains independent: it may
stop an action even after evidence verification and authorized finalization.

This paper makes four contributions. First, it defines the verification gap as
a communication-relevant state between proposal understanding and physical
action. Second, it proposes the Post-Semantic Communication Framework, which
connects evidence requirements, evidence records, verification, authorized
finalization, and a downstream runtime boundary. Third, it implements the
framework in a finite-state evaluator and checks consistency with the declared
structural distinctions,
including conflict-first verification and the separation between evidence
transfer and evidence coordination. Fourth, a controlled communication study
demonstrates a requirement-preserving comparison and supplies reporting
guidelines and an episode-level reporting schema for reuse. The checks stop at
authorized finalization; measured sensing, credential operation, runtime
control, and physical safety remain future work.

\section{Related Work and Motivation}

Semantic and task-oriented communication move the objective beyond symbol
fidelity toward meaning and task effectiveness \cite{gunduz,pragcomm}. Our
question begins after a meaningful proposal already exists. In Weaver's broad
effectiveness sense, the framework isolates the post-proposal transition from
task effectiveness to evidence-backed, authorized action toward Level~C
\cite{weaver}; it does not redefine communication
theory. Receiver side information and interactive coding explain why a sender
need not retransmit what a receiver already has \cite{wynerziv,kaspi}, while
learned multi-agent communication asks when agents should exchange information
\cite{when2com}. These lines motivate efficient delivery, but do not by
themselves declare which evidence must be verified before a particular
physical action is finalized.

Active perception and information-directed sensing ask what observation to
acquire next \cite{bajcsy,mackay}; distributed sensor fusion asks how
heterogeneous observations should be combined \cite{hallllinas}. Distributed
detection, distributed hypothesis testing, and team decision theory show how
observations and decisions can be distributed \cite{tenney,ahlswede}. Work on
information structures in distributed control similarly shows that who knows
what, and when, constrains achievable decisions. The present framework fixes
the required evidence and authorized finalizer before comparing how already
available records reach that endpoint. Physical evidence location is therefore
not silently credited as authority over an action.

Embodied multimodal models ground sensors in language, while
closed-loop LLM planners use scene descriptions, success signals, and human
feedback to replan \cite{palme,innermonologue}. Uncertainty-aware planners can
ask supervisors for help under ambiguity, while interactive multimodal systems
can choose epistemic actions to gather visual, acoustic, haptic, or
proprioceptive observations \cite{knowno,chatenv}. They motivate AI-assisted
diagnosis and information seeking, but do not themselves define which
observation qualifies as action-specific evidence, how it is bound to a
proposal, or who may authorize finalization. The proposed framework supplies
that verification and authority boundary.

Selective prediction supplies Accept, Reject, and Abstain outcomes
\cite{selectivenet}. Provenance and verifiable-sensing work motivate checkable
records \cite{prov,proofsensing,verifiablesemantics}; safety and assurance cases
motivate explicit claim--evidence structure \cite{gsn3}; runtime assurance
protects execution after a decision \cite{rta}; and zero-trust and
capability-based mechanisms inform delegated command authorization
\cite{nistzt,macaroons}. These mechanisms address different layers. A valid
record does not grant action authority, and authorized finalization does not
replace current-state runtime enforcement.

Contract-based design and assume--guarantee reasoning associate components
with environmental assumptions and behavioral guarantees, supporting
compositional design, refinement, and verification
\cite{benvenistecontracts}. Such contracts support compositional verification
of modular robots and ROS nodes \cite{luckcuckrobotics}. Our evidence contract
instead declares action-specific claims and validity conditions for
proposal-bound records at the permitted finalizer after proposal formation; it
does not specify component behavior under environmental assumptions. They are
complementary: component contracts constrain behavior; an evidence contract
governs what information may justify an action.

Our contribution is to make the post-proposal state an explicit communication
framework. It binds application-declared evidence requirements to
proposal-specific records, separates verification from authorized finalization
and runtime enforcement, and exposes evidence transfer and evidence
coordination as distinct communication functions. Related fields contain
individual elements of this design; the framework integrates them around the
verification gap and a common episode boundary.

\begin{table*}[!t]
\caption{Same Proposal, Different Action Outcomes.}
\label{tab:counterexamples}
\centering
\small
\begin{tabularx}{0.96\textwidth}{@{}L{0.15\textwidth}L{0.27\textwidth}L{0.29\textwidth}X@{}}
\toprule
Changed condition & Proposal-only view & Evidence-aware view & Required response \\
\midrule
Proposal binding & ``Proceed'' is unchanged. & The record hash refers to an
earlier proposal. & Abstain: the record cannot satisfy the current requirements. \\
Freshness / expiry & ``Proceed'' is unchanged. & A required record expires
before finalization. & Abstain: evidence is no longer timely. \\
Admissible conflict & ``Proceed'' is unchanged. & A validated record conflicts
with a required claim. & Reject before testing support completeness. \\
Finalization authority & ``Proceed'' is unchanged. & The endpoint holding
sufficient records is not authorized to finalize. & Make the required evidence
available to the authorized finalizer, or abstain. \\
\bottomrule
\end{tabularx}
\end{table*}

Table~\ref{tab:counterexamples} is a representational counterexample, not an
impossibility theorem: a richer proposal could carry these fields. The point is
to preserve them as explicit, independently reportable framework objects.

\section{Post-Semantic Communication Framework}

\begin{figure*}[!t]
\centering
\includegraphics[width=0.98\textwidth]{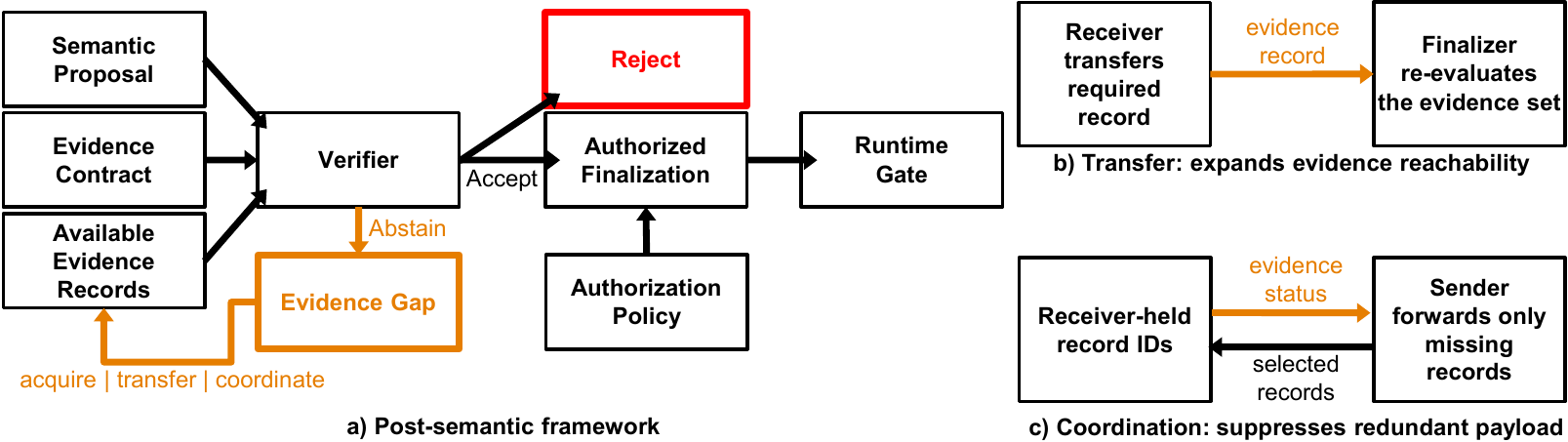}
\caption{The Post-Semantic Communication Framework. (a) A semantic proposal, an application-declared evidence contract, and available evidence records enter a common verifier. An admissible conflict yields Reject; complete support with no admissible conflict yields Accept, which can be finalized only under the separate authorization policy; missing, invalid, stale, or incomplete evidence yields Abstain and an evidence gap. The gap may trigger acquisition, transfer, or coordination followed by re-verification, while the runtime gate remains downstream of authorized finalization and outside the present evaluation. (b) Evidence transfer moves a required record to the finalizer, expanding evidence reachability. (c) Evidence coordination exchanges record-status information so that the sender forwards only missing records, suppressing redundant payload; status is not evidence. Acquisition is a future extension, whereas Section IV evaluates transfer and coordination.}
\label{fig:framework-architecture}
\end{figure*}

Section~II identified episode distinctions that a proposal-only representation
can hide. We now define the proposed framework that makes those distinctions
explicit and communicable; this is the paper's main contribution. The framework
answers four questions: what evidence is required; how records are validated
and combined; who may finalize and what remains for runtime control; and how
communication can transfer evidence or coordinate delivery.
Figure~\ref{fig:framework-architecture} summarizes the resulting architecture.

\subsection{Evidence Requirements and Records}

An \emph{evidence contract} is the application-declared set of evidence
requirements and validity rules that must be satisfied before an action can be
finalized. It specifies the required claims, acceptable record modalities and
sources, support and conflict semantics, freshness and uncertainty limits,
provenance checks, proposal binding, and evidence deadline. We use
\emph{evidence requirements} for the individual conditions and \emph{evidence
contract} for the formal collection of those conditions. A communication
policy may change delivery but must not silently relax this contract.

An \emph{evidence record} identifies its claim, relation to that claim,
modality, source, measurement time, freshness limit, provenance, uncertainty,
proposal binding, and payload or content hash. These fields state which
contract item the record can satisfy and allow a verifier to check the record
mechanically. They do not define a new data type: an image, radio observation,
signed state estimate, or reference can be an evidence record when it carries
the required attributes. In the warehouse example, one record addresses
corridor clearance and another addresses the absence of a moving obstacle.

Proposal binding prevents a record for an earlier route or control cycle from
satisfying the current requirements. Decision-time freshness prevents a record
that was valid when transmission began from being accepted after expiry.
Provenance, source, calibration, and uncertainty rules determine whether the
application admits the record. A delivery policy may change packetization,
routing, references, or interaction without changing any of these tests.

\subsection{Evidence Verification}

Supporting and conflicting records traverse the same validation pipeline.
Expired, untrusted, calibration-revoked, excessively uncertain, or
binding-mismatched records are inadmissible: they can neither satisfy an
evidence requirement nor trigger \emph{Reject}. This common path prevents an invalid
negative observation from bypassing the checks applied to support.

The verifier follows the principle \emph{Conflict First, Then Sufficiency}.
Among admissible records, any conflict produces \emph{Reject}; a complete
supporting set with no conflict produces \emph{Accept}; and every remaining
case produces \emph{Abstain}. Reject reports a contradiction. Abstain reports
that the declared evidence requirements were not completed. A conditional
record error can still create incorrect support, so verification establishes
compliance with the declared contract rather than physical truth.

An unresolved verification gap takes one of two forms: an \emph{evidence gap},
when required admissible records are missing, invalid, stale, incomplete, or
unavailable at the permitted finalizer; or an \emph{authority gap}, when
evidence is sufficient but authorized finalization is unavailable. An
admissible conflict is not a gap: it is a resolved negative outcome that
produces \emph{Reject}. An inadmissible conflict does not enter the verifier's
admissible set and therefore cannot trigger \emph{Reject}.

\subsection{Finalization and Runtime Safety}

Evidence sufficiency, authorized finalization, and runtime safety are three
separate responsibilities. Evidence sufficiency is the verifier's result over
admissible records. An \emph{authorization policy} is a separate application
declaration that identifies the endpoint or endpoint set permitted to turn an
Accept result into an accountable action decision. Authorized finalization
requires both verifier Accept and membership in that declared set; authority is
not part of the evidence contract. Runtime safety is the later decision of an
independent gate using current physical conditions.

An agent can contribute a record without acquiring permission to act. An
authorized finalizer can use remote records without claiming to have measured
them locally. If sufficient records are assembled at an unauthorized endpoint,
the records must become available to the declared finalizer or the system must
abstain. A content-addressed manifest may preserve record-set identity across
that handoff, but it creates neither authority nor freshness.

The runtime gate receives a finalized action and may return ACT or STOP. It can
account for conditions not represented in the evidence contract, including a
new obstacle or degraded actuator. Section~IV evaluates only through authorized
finalization. It does not evaluate runtime-gate performance, sensor-statement
correctness, operational credential enforcement, or physical safety.

\subsection{Evidence Transfer and Coordination}

Post-semantic communication can serve two functions. \emph{Evidence transfer}
carries an admissible record that the authorized finalizer does not yet hold.
For sender finalization, successful feedback may return a receiver-held record
and thereby enlarge the admissible set reachable before finalization. This
distinction gives Feedback a less obvious role: even when the sender retains
finalization authority, the reverse message may carry missing evidence back to
the decision point rather than merely report receiver state. In contrast,
\emph{evidence coordination} reports which records are already held at the
receiver so that the sender can avoid transmitting redundant payload. When the
receiver finalizes, Feedback instead coordinates delivery around records
already available at the decision point. The status message guides delivery but
does not satisfy an evidence requirement.

The controlled evaluator compares the formal policies \emph{Sender/One-way},
\emph{Sender/Feedback}, \emph{Receiver/One-way}, and
\emph{Receiver/Feedback}. The first term fixes the authorized finalizer; the
second changes interaction. A One-way policy sends all validated sender-held
records. Feedback adds a reverse logical message. Under receiver finalization,
loss of status triggers the declared One-way fallback, so coordination does
not acquire an artificial reachability advantage.

Both functions have costs. A reverse message can be lost, add fixed latency,
age a record past its lifetime, or exhaust the decision deadline. Conversely,
a short status exchange can suppress a large redundant record. These are
communication effects; changes in the evidence contract, finalizer, or runtime
gate must be evaluated separately.

The framework therefore imposes four comparison invariants: evidence
requirements remain fixed within a communication comparison; supporting and
conflicting records share one validation boundary; evidence location does not
confer finalization authority; and runtime outcomes are not credited to
pre-action verification. Section~V-B carries these invariants into the
reporting checklist.

\section{Framework Checks and Communication Study}

\subsection{Questions and Setup}

This section checks the framework's internal distinctions and implementation
consistency, and then studies communication choices under controlled
assumptions. It does not evaluate deployment effectiveness or runtime safety.
The analysis asks three questions. \emph{RQ1:} Can the framework distinguish
same-proposal episodes that differ in proposal binding, freshness, conflict, or
finalization authority? \emph{RQ2:} Does the evaluator expose evidence transfer and evidence
coordination as distinct observable effects? \emph{RQ3:} Can One-way and
Feedback be compared while evidence requirements, finalizer, and evidence
quality are held fixed?

The synthetic setup has two evidence items and two endpoints. Each endpoint
observation is missing, invalid, supporting, or conflicting; forward and
reverse delivery each succeed or fail. The same 4096 weighted
states are enumerated once and reused across all four communication cases.
Records follow the common validation path, freshness is rechecked at decision
time, and the finalizer is fixed within each comparison. The episode begins
after proposal formation and ends at authorized finalization. The common 10-ms
processing time is included, and the baseline deadline is 33~ms. Section~IV-B
operationalizes the cases motivated by Table~\ref{tab:counterexamples}.

Unconditional coverage is the probability of timely Accept over all proposal
episodes. Selective error is the ground-truth-unsupported mass among timely
accepts. Expected evidence payload counts transmitted evidence-record bytes
over all proposal episodes, including episodes that later Reject or Abstain.
These metrics describe the declared finite model; they are not measured robot,
wireless, or safety results. Headers, acknowledgments, retransmissions, channel
occupancy, airtime, and energy are not represented by the payload field.

\subsection{Framework Checks}

\emph{RQ1 -- structural distinctions.} The artifact executes
11 checks and passes all
11. Complete admissible support with no conflict
returns Accept; an admissible conflict returns Reject even before support is
complete; and incomplete support returns Abstain. Expired, untrusted,
binding-mismatched, calibration-revoked, and excessively uncertain conflicts
do not trigger Reject. Finally, complete evidence at an unauthorized endpoint does not
produce authorized finalization. These checks operationalize the distinctions
in Table~\ref{tab:counterexamples}; they test declared framework logic, not the
physical correctness of a sensor claim.

\emph{RQ2 -- transfer versus coordination.} With timing and expiry removed for
this structural check, sender-finalized Feedback uniquely reaches a complete
admissible support set in 64 enumerated states.
Under the declared synthetic state distribution, these states carry
8.3\% probability mass. This is an
evidence-transfer effect, not an estimate of real episode frequency. Under
receiver finalization and the declared One-way fallback, the pre-timing
reachability mismatch between One-way and Feedback is
0 states. Feedback instead reduces expected
evidence payload by 6.49~KiB per proposal episode,
an evidence-coordination effect without enlarging the reachable evidence set;
its extra message can nevertheless change the timely result. At the ordinary
timing baseline, the generated expected payloads are
9.80~KiB for One-way,
8.45~KiB for sender-finalized Feedback, and
3.31~KiB for receiver-finalized Feedback; the
matched receiver-finalized One-way value is 9.80~KiB.
These checks isolate the two mechanisms that Figure~\ref{fig:requirement-regime}
later combines with timing: transfer changes the reachable evidence set at a
sender finalizer, whereas coordination leaves pre-timing reachability unchanged
under the declared fallback but reduces payload at a receiver finalizer.

\subsection{Communication Study}

\begin{figure*}[!t]
\centering
\includegraphics[width=0.98\textwidth]{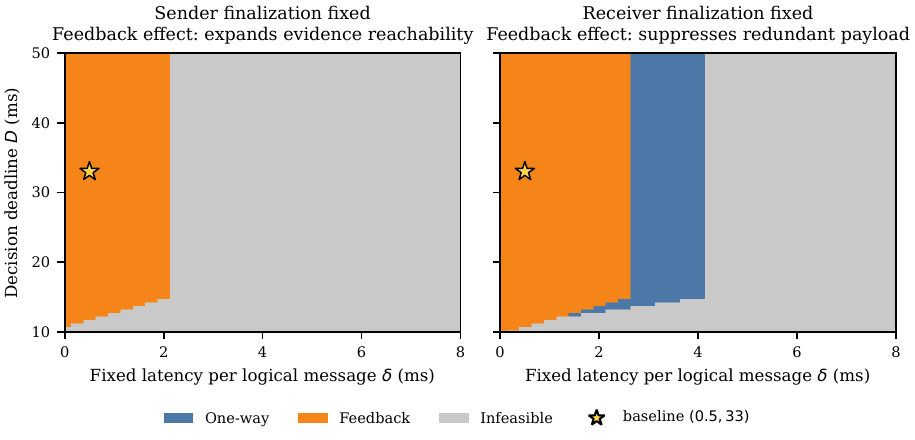}
\caption{Controlled communication study under the declared synthetic model,
with evidence requirements and evidence-quality assumptions fixed across cases
and finalization authority fixed within each panel. Orange, blue, and gray
denote Feedback selected, One-way selected, and infeasible, respectively. Every
feasible sender-finalized cell is orange: reverse transfer can make
receiver-held evidence available at the sender decision point, while One-way
violates the selective-error requirement in those cells; tighter timing makes
both alternatives infeasible. For receiver finalization, Feedback is selected
at low latency, One-way when the extra reverse message makes Feedback miss the
coverage requirement, and both eventually become infeasible. Feasibility
requires at least 40\% unconditional coverage and at most 0.75\% selective
error; the star marks the baseline.}
\label{fig:requirement-regime}
\end{figure*}

\emph{RQ3 -- requirement-preserving communication choice.} RQ3 adds message
latency, record ageing, and the decision deadline to the two structural effects
isolated in RQ2. Figure~\ref{fig:requirement-regime}
holds the two evidence requirements, authorized finalizer, evidence-quality
model, record sizes, link rate, processing time, initial ages, and lifetime
fixed. It varies only fixed latency per logical message and the decision
deadline. A candidate is feasible only if unconditional coverage is at least
40\% and selective error is at most 0.75\%. Among feasible candidates, the
declared rule minimizes expected evidence payload, then logical-message count,
with an exact tie assigned to One-way. The resulting map contrasts the two
communication benefits with their timing costs.

The orange region denotes cells in which Feedback is selected; it does not
imply a single selection reason. The sender-finalized panel exposes the less
obvious role of Feedback. Although the sender retains finalization authority,
all 675 feasible plotted cells select Feedback under the declared model. In
each, Feedback is the only feasible candidate because One-way exceeds the
selective-error ceiling. Feedback can transfer receiver-held evidence to the
sender decision point, whereas One-way can use only sender-held evidence there.
As latency rises or the deadline tightens, both alternatives fail the declared
feasibility requirements in the other 1,998 cells; no sender-finalized cell
selects One-way on the plotted grid.

The receiver-finalized panel shows the complementary role. Feedback does not
enlarge pre-timing evidence reachability under the declared fallback; it uses
receiver-held record status to suppress redundant forward payload. Of the 836
Feedback-selected cells, both alternatives are feasible in 830 and Feedback
minimizes expected payload; in the other 6, only Feedback is feasible. The 441
One-way-selected cells occur when the extra reverse message makes Feedback miss
the coverage requirement. In the remaining 1,396 cells, both alternatives miss
coverage and are infeasible.

Where finalization occurs determines what Feedback is for: bringing missing
evidence back to a sender finalizer or avoiding redundant evidence delivery to
a receiver finalizer. This finalizer-dependent selection structure is specific
to the declared synthetic model and plotted grid and does not establish a
protocol ranking beyond those conditions.

\section{Discussion and Future Directions}

\subsection{Research Directions}

\emph{Evidence-gap detection} separates deterministic checking from AI
assistance. Contract checks identify missing, invalid, expired, or
binding-mismatched records; authorization checks find an unavailable finalizer;
admissible conflict yields Reject. Embodied multimodal models integrate
language and sensor inputs \cite{palme}; uncertainty-aware planners seek help
under ambiguity \cite{knowno}. Within our framework, they could propose evidence
needs or clarifications; checks still distinguish evidence gaps, authority
gaps, and conflicts. Neither confidence nor diagnosis is an evidence record
or replaces checks.

\emph{Active acquisition and representation} maps camera, LiDAR, contact, peer,
human, or radio sensing to named requirements under freshness, uncertainty, and
deadlines. Interactive agents select visual, acoustic, haptic, and
proprioceptive information-gathering actions \cite{chatenv}; ISAC and perceptive
mobile networks jointly support radio communication and environmental sensing
\cite{rahmanperceptive,liu}. Within our framework, observations qualify as
evidence only after requirement mapping and validation of proposal binding,
provenance, freshness, uncertainty, and source. Semantic and task-oriented
communication support task-relevant representation and delivery
\cite{gunduz,pragcomm}. The representation must additionally preserve the claim
relation and those validation attributes.

\emph{Closed-loop re-inference and runtime integration} could build on embodied
multimodal models, LLM planners, and world-model or digital-twin architectures
that exemplify reasoning, replanning, and state-dependent revision
\cite{palme,innermonologue,wmcdt}. Within our framework, such engines could use
validated evidence records to revise a proposal. Revision creates a new
binding, may select another application-approved template, and requires
revalidation. The runtime gate remains downstream and unevaluated. AI agents
may diagnose and plan, but the evidence contract, common validator,
authorization policy, and runtime gate remain explicit decision boundaries.

\subsection{Reporting Guidelines}

Table~\ref{tab:checklist} gives the minimum fields needed to preserve the
separations in Section~III when comparing communication designs.

\begin{table*}[!t]
\caption{Minimum Reporting Checklist for Post-Semantic Comparisons.}
\label{tab:checklist}
\centering
\small
\begin{tabularx}{0.95\textwidth}{@{}L{0.16\textwidth}L{0.37\textwidth}X@{}}
\toprule
Layer & Report & Confounding avoided \\
\midrule
Evidence & Requirements; support/conflict semantics; record source, modality,
freshness, provenance, uncertainty, and proposal binding. & A change in what
counts as sufficient is not mislabeled as a delivery gain. \\
Delivery & One-way or Feedback; feedback function; success model; expected
evidence payload per proposal episode; logical-message count; latency and
deadline accounting. & Payload suppression is not confused with total on-wire
or energy reduction. \\
Finalization & Authorized finalizer; evidence-assembly location; credential and
handoff assumptions. & Evidence location is not mistaken for action authority. \\
Runtime & Gate inputs, re-evaluation time, stop or backup action, and the
execution-observation boundary. & Verifier Accept is not presented as a
physical-safety result. \\
\bottomrule
\end{tabularx}
\end{table*}

First, each field should be declared before comparison as fixed, varied, or
unobserved. Evidence requirements, finalizer, freshness, and deadline are
application inputs; the communication policy is the experimental factor.

Second, an episode-level trace should link the proposal, admissible and rejected
records, messages, verifier outcome, authorized finalizer, and any observed
runtime result. Stable record identifiers and validation outcomes can preserve
this structure without publishing sensitive payloads.

Third, aggregate metrics should use explicit and common denominators.
Unconditional coverage and expected payload use all proposal episodes;
selective error uses timely Accept episodes; runtime rates use only episodes in
which execution was observed. This accounting keeps the framework layers
comparable without inventing unmeasured outcomes.

\section{Conclusion}

A task-effective proposal is not yet a justified physical action. The
verification gap is the mismatch between such a proposal and the valid, timely,
proposal-bound evidence and authorization required at the permitted finalizer
before action. Post-semantic communication makes explicit what must happen after
a meaningful proposal is formed but before a physical action may proceed. The
Post-Semantic Communication Framework makes the missing objects and
responsibilities explicit through evidence requirements, proposal-bound
records, conflict-first verification, authorized finalization, and a downstream
runtime boundary. Where finalization occurs determines what Feedback is for:
evidence transfer brings missing evidence to a sender finalizer, whereas
evidence coordination avoids redundant evidence delivery around records already
held at a receiver finalizer. The finite-state
checks verify implementation consistency with the declared operational
distinctions, while the controlled communication study demonstrates one
requirement-preserving comparison of One-way and Feedback. The reporting
guidelines carry the same separations into future
studies. Runtime effectiveness and physical safety remain subjects for measured
systems; runtime control stays downstream of verification and authorized
finalization.

\appendices

\section{Formal Observations and Structural Properties}

\emph{Observation 1 --- Proposal-only representations can collapse distinct
outcomes.} Let $\pi(w)$ be a reduced proposal representation and let
$a^\star(w)$ be the required outcome. Suppose two physical or system states
satisfy
\begin{equation}
\pi(w_1)=\pi(w_2),\qquad a^\star(w_1)\ne a^\star(w_2).
\end{equation}
If the outcome difference is caused by discarded proposal binding,
decision-time freshness, admissible conflict, or finalization authority, no
decision rule based only on $\pi(w)$ can be correct in both states.

\emph{Justification.} For a proposal-only rule $g$, equality of the reduced
representations gives
\begin{equation}
g(\pi(w_1))=g(\pi(w_2)).
\end{equation}
The required outcomes differ, so the common value must disagree with at least
one state. This is a statement about the reduced representation; a richer
semantic representation could carry the omitted fields. \hfill$\square$

\emph{Structural Property 1 --- Conflict-first verifier consistency.} Let
$V_q(e,d,t)$
be the common validation predicate for evidence record $e$, proposal $d$,
decision time $t$, and contract $q$. Define the admissible set as
\begin{equation}
E_q(d,t)=\{e:V_q(e,d,t)=1\}.
\end{equation}
The verifier
\begin{equation}
\Gamma_q(d,E_q)=
\begin{cases}
\mathrm{Reject}, & \text{an admissible conflict exists},\\
\mathrm{Accept}, & \text{required support is complete},\\
\mathrm{Abstain},& \text{otherwise}
\end{cases}
\end{equation}
satisfies three properties: inadmissible records neither contribute support nor
trigger Reject;
admissible conflict precedes support completeness; and complete admissible
support without conflict is accepted.

\emph{Argument.} Validation first removes every record for which $V_q=0$, establishing
the first property. The first branch handles every remaining conflict before
the support branch. If that branch is false and the required support is
complete, the second branch applies; all residual cases enter the third branch.
These exhaustive cases establish consistency with the declared verifier rule.
They do not establish physical safety or deployment correctness.
\hfill$\square$

Authorized finalization is a separate predicate. For endpoint $i$ and declared
authorization policy $\mathcal{A}$, an Accept result can authorize action only
when
\begin{equation}
\Gamma_q=\mathrm{Accept}\quad\text{and}\quad i\in\mathcal{A}.
\end{equation}

\section{Communication, Freshness, and Availability Model}
\label{app:timing}

Fixed latency $\delta$ is charged once per logical message, evidence payload is
serialized at rate $R$, and common pre-action processing $T_p$ covers
validation, assembly, and finalization. The ready times are
\begin{align}
T_{1}(B_f)&=\delta+8B_f/R+T_p,\\
T_{F}(B_r,B_f)&=2\delta+8(B_r+B_f)/R+T_p.
\end{align}
For initial record age $A_e$, lifetime $L_e$, and deadline $D$, a record used at
finalization must satisfy
\begin{equation}
A_e+T_x\leq L_e,\qquad T_x\leq D.
\end{equation}

For one evidence item, let endpoint holding indicators have marginal
probabilities $q_S$ and $q_R$ and Pearson correlation $\rho$. The evaluator
uses
\begin{equation}
\Pr(H_S=1,H_R=1)=q_Sq_R+
\rho\sqrt{q_S(1-q_S)q_R(1-q_R)}.
\end{equation}
The joint probability must lie within the Fr\'echet bounds; the remaining three
Bernoulli cells follow from the marginals. This correlation applies across
endpoints for each item, not between the two independent claim truths.

\section{Feedback Function Separation and Requirement-Based Selection}

\emph{Structural Property 2 --- Separation of feedback functions.} Consider a fixed evidence
contract and finalizer. For sender finalization, if a receiver holds a missing
admissible record, the reverse transfer succeeds, and the record remains valid
at decision time, Feedback can strictly enlarge the admissible evidence set
reachable at the finalizer. For receiver finalization under a One-way fallback,
coordination does not enlarge pre-timing evidence reachability relative to
One-way, but successful status delivery can weakly reduce forward evidence
payload when the receiver already holds a required record. Added latency or
expiry may still change the timely outcome.

\emph{Argument.} In the first case, let $E_S$ be the sender-held admissible set and
$e_R$ a required record held only by the receiver. Successful transfer yields
\begin{equation}
E_S\subset E_S\cup\{e_R\},
\end{equation}
with strict inclusion; the added record may complete support. In the second
case, One-way sends every sender-held admissible record. Delivered status lets
the sender omit only records already available at the receiver; lost status
invokes the same One-way set. The receiver's pre-timing union is therefore
unchanged, while transmitted payload is no larger and is smaller whenever a
positive-size redundant record is omitted. Feedback has an extra logical
message, so decision-time freshness and deadline tests can reverse the timely
outcome. \hfill$\square$

For fixed finalization authority, let $C_x$, $E_x$, $B_x$, and $N_x$ denote
unconditional coverage, selective error, expected evidence payload per proposal
episode, and logical-message count for interaction $x$. Feasibility requires
\begin{equation}
C_x\geq C_{\min},\qquad E_x\leq E_{\max}.
\end{equation}
Among feasible candidates, the controlled illustration minimizes
\begin{equation}
(B_x,N_x,I_x)
\end{equation}
lexicographically, where $I_x$ resolves an exact tie in favor of One-way. This
is an application-declared ordering rather than a general utility. Sender-finalized
Feedback can improve $C_x$ through transfer; receiver-finalized Feedback can
reduce $B_x$ through coordination. In either case, an added message can reduce
timely coverage through reverse loss, latency, record ageing, or expiry.

\section{Synthetic Setup and Reproduction Boundary}

Table~\ref{tab:baseline-parameters} records the fixed synthetic declarations.
The per-claim truth prior and per-endpoint evidence availability happen to be
0.90 but are different quantities. The two truth bits are independent;
availability for the same item is correlated across endpoints as defined in
Appendix~\ref{app:timing}.

\begin{table}[H]
\caption{Synthetic Baseline Declarations.}
\label{tab:baseline-parameters}
\centering
\footnotesize
\begin{tabular}{@{}ll@{}}
\toprule
Parameter & Baseline \\
\midrule
Required evidence & Visual clearance + Radio no-motion \\
Claim-truth prior $p_{\mathrm{truth}}$ & 0.90 per claim \\
Claim-truth dependence & Independent \\
Evidence availability $q$ & 0.90 per endpoint/item \\
Conditional validity & 0.99 \\
Evidence-relation error & 0.10 \\
Endpoint correlation $\rho$ & 0.25 \\
Delivery success $(p_f,p_r)$ & (0.99, 0.99) \\
Records $(B_V,B_R)$ & (10, 1) KiB \\
Link rate $R$ & 100 Mbit/s \\
Per-message latency $\delta$ & 0.5 ms \\
Processing time $T_p$ & 10 ms \\
Decision deadline $D$ & 33 ms \\
Initial ages $(A_V,A_R)$ & (20, 15) ms \\
Evidence TTL & 35 ms \\
\bottomrule
\end{tabular}
\end{table}

Exact enumeration weights two truth bits, four endpoint-record observations,
and two delivery bits. Missing, invalid, supporting, and conflicting
observations share the same weighted 4096-state population in
all four communication cases. Communication selects validated records, charges
their bytes, computes action-ready time, revalidates lifetime, and returns
Accept, Reject, or Abstain; a separate authorization check determines whether
Accept can be finalized. Enumeration removes Monte Carlo sampling noise only.

The 11 generated RQ1 checks cover complete and
incomplete support, an admissible Visual conflict, conflict before support
completeness, six conflicts made inadmissible by provenance, binding, expiry,
calibration, source, or uncertainty, and complete evidence at an unauthorized
endpoint. They exercise Accept, Abstain, and conflict-first Reject, verify that
inadmissible conflicts do not trigger Reject, and separate evidence sufficiency
from the authorization policy.

RQ2 uses the same weighted states without deadline or lifetime truncation. The
sender-transfer metric measures states in which Feedback reaches complete
support and One-way does not. The receiver metric checks One-way/Feedback
reachability under the declared status-loss fallback, while an unconditional
expectation measures evidence-record bytes suppressed by coordination. Timing
is restored for the baseline values and Figure~\ref{fig:requirement-regime}.

For RQ3, every grid cell reuses the same truth, record-observation, and delivery
population. Only fixed latency per logical message and the deadline vary. The
generator recomputes timely Accept and selective error, applies the common
feasibility requirements, and then the declared payload/message ordering.
Generated LaTeX values and validation preserve source-to-claim consistency
within the declared finite model.

The model contains no measured wireless trace, queue, scheduler, retransmission
model, calibrated sensor dataset, exercised authorization infrastructure, or
measured runtime controller. Figure~\ref{fig:requirement-regime} is therefore a
constructed design result rather than deployment evidence, and the runtime gate
is only a downstream boundary.

The code and generated artifacts used for this manuscript version are
available in the public Technical Companion at
\url{https://github.com/sarulab-ou/post-semcom}. They establish internal
consistency, not calibration, certification, sensor correctness, runtime
effectiveness, or physical safety.

\bibliographystyle{IEEEtran}
\bibliography{reference}